# Region-Based Approximations for Planning in Stochastic Domains

**Nevin L. Zhang and Wenju Liu**
Department of Computer Science
Hong Kong University of Science and Technology
{lzhang, wliu}@cs.ust.hk

## Abstract

This paper is concerned with planning in stochastic domains by means of partially observable Markov decision processes (POMDPs). POMDPs are difficult to solve. This paper identifies a subclass of POMDPs called region observable POMDPs, which are easier to solve and can be used to approximate general POMDPs to arbitrary accuracy.

## 1 INTRODUCTION

To plan is to find a policy that will lead an agent to achieve a goal with minimum cost. When the environment of the agent, henceforth referred to as the world, is completely observable and the effects of actions are deterministic, planning is reduced to finding the shortest sequence of actions that leads the agent to the goal.

In real-world applications, however, the world is rarely completely observable and effects of actions are almost always nondeterministic. For this reason, a growing number of researchers concern themselves with planning in stochastic domains (e.g. Dean and Wellman 1991, Cassandra *et al* 1994, Boutillier *et al* 1995, Parr and Russell 1995). Partially observable Markov decision processes (POMDPs) can be used as a model for planning in such domains. In this model, nondeterminism in effects of actions is encoded by transition probabilities, partial observability of the world by observation probabilities, and goals and criteria for good plans by reward functions.

POMDPs are difficult to solve and approximation is a must in real-world applications. Most previous approximation methods (e.g. Cheng 1988, Lovejoy 1991b, and Parr and Russell 1995) are *value function approximation methods* in the sense that they approximate optimal value functions of POMDPs directly. We advocate *model approximation methods*. Such a method approximates a POMDP itself by another that is easier to solve and uses the solution of the latter to construct an approximate solution to the original POMDP.

Model approximation can be in the form of a more informative observation model, or a more deterministic action model, or an aggregation of the state space, or a combination of two or all of them. This paper investigates the first alternative.

The idea of approximating a POMDP by assuming a more informative observation model is not new. Cassandra *et al* (1996) have proposed to approximate POMDPs by using MDPs. This paper generalizes the idea. We transform a POMDP by assuming that, in addition to the observations obtained by itself, the agent also receives a report from an oracle who knows the true state of the world. The oracle does not report the true state itself. Rather, he selects, from a list of candidate regions, a region that contains the true state and reports that region. The transformed POMDP is said to be *region observable* because the agent knows for sure that the true state is in region reported by the oracle.

When all candidate regions are singletons, the oracle actually reports the true state of the world. In such a case, the region observable POMDP reduces to an MDP. MDPs are much easier to solve than POMDPs. One would expect the region observable POMDP to be solvable when all candidate regions are small.

In terms of quality of approximation, the larger the candidate regions, the less extra information the oracle provides and hence the more accurate the approximation. In the extreme case when there is only one candidate region and it consists of all possible states of the world, the oracle provides no extra information at all. Hence the region observable POMDP is identical to the original POMDP.

A way to determine the quality of approximation will be described. This allows one to make the tradeoff between approximation quality and computational complexity as follows: start with small candidate regions and increase their sizes gradually until the approximation becomes accurate enough or the region observable POMDP becomes untractable.

In many applications, the agent often has a good idea about the true state of the world. Take robot path planning as an example. Observing a landmark, a room number for instance, would imply that the robot is at the proximity of that landmark. Observing a feature about the world, a corridor T-junction for instance, might imply the robot is in one of several regions. Taking history into account, the robot might be able to determine a unique region for its current location. Also, an action usually moves the true state of the world to only a few "nearby" states. Thus if the robot has a good idea about the current state of world, it should continue to have a good idea about it in the next few steps.

When the agent has a good idea about the true state at all time, accurate approximation can be achieved with small candidate regions.

We shall begin with a brief review of planning under uncertainty and POMDPs. We shall then formally introduce region observable POMDPs as an approximation to general POMDPs. Thereafter, we shall describe a way to determine the quality of approximation. Finally, we shall report empirical results, which suggest that when there is not much uncertainty, a POMDP can be approximated accurately by a region observable POMDP that has small candidate regions and can hence be solved exactly.

## 2 PLANNING UNDER UNCERTAINTY AND POMDPs

To specify a planning problem, one needs to give a set $S$ of possible states of the world, a set $O$ of possible observations, and a set $A$ of possible actions. In this paper, all those three sets are assumed to be finite. One needs also to give an observation model, which describes the relationship between an observation and the state of the world; and an action model, which describes the effects of each action. Furthermore, one needs to specify the initial state of the world and a goal state.

As a background example, consider path planning for a robot who acts in an office environment. Here $S$ is the set of all location-orientation pairs, $O$ is the set of possible sensor readings, and $A$ consists of actions move-forward, turn-left, turn-right, and declare-goal.

The current observation $o$ depends on the current state of the world $s$. Due to sensor noise, this dependency is uncertain in nature. The observation $o$ sometimes also depends on the action that the robot has just taken $a_-$. The minus sign in the subscript indicates the previous time point. In the POMDP model, the dependency of $o$ upon $s$ and $a_-$ is numerically characterized by a conditional probability $P(o|s, a_-)$, which is usually referred to as the *observation probability*. It is the observation model.

In a region observable POMDP, the current observation also depends on the previous state of the world $s_-$. The observation probability for this case can be written $P(o|s, a_-, s_-)$.

The state $s_+$ the world will be in after taking an action $a$ depends on the action and on the current state $s$. The plus sign in the subscript indicates the next time point. This dependency is again uncertain in nature due to uncertainty in the actuator. In the POMDP model, the dependency of $s_+$ upon $s$ and $a$ is numerically characterized by a conditional probability $P(s_+|s, a)$, which is usually referred to as the *transition probability*. It is the action model.

We will often need to consider the joint conditional probability $P(s_+, o_+|s, a)$ of the next state of the world and the next observation given the current state and the current action. It is given by

$$P(s_+, o_+|s, a) = P(s_+|s, a)P(o_+|s_+, a, s).$$

The POMDP model encodes the starting state by a probability mass function $P_0$ over $S$. The planning goal is encoded by a *reward function* such as the following:

$$r(s,a) = \begin{cases} 1 & \text{if } a\text{=delcare-goal and } s\text{=goal,} \\ 0 & \text{otherwise.} \end{cases} \quad (1)$$

## 3 DECISION MAKING IN POMDPs

The agent chooses and executes an action at each time point. The choice is made based on the agent's knowledge about the true state of the world, which is summarized by a probability distribution over the set of possible states and called a *belief state*. The initial belief state is $P_0$. Suppose $b$ is the current belief state, and $a$ is the current action. If the observation $o_+$ is obtained at the next time point, then the next belief state $b_+$ is given by

$$b_+(s_+) = k \sum_s P(s_+, o_+|s, a)b(s), \quad (2)$$

where $k=1/\sum_{s,s_+} P(s_+, o_+|s, a)b(s)$ is the normalization constant (Cassandra et al 1994). To signify the dependence of $b_+$ upon $b$, $a$, and $o_+$, we shall sometimes write it as $b_+(.|b, a, o_+)$.

A *policy* $\pi$ prescribes an action for each possible belief state. Formally it is a mapping from the set $\mathcal{B}$ of all possible belief states to $\mathcal{A}$. For each belief state $b$, $\pi(b)$ is the action prescribed by $\pi$ for $b$. The *value function* of $\pi$ is defined for all belief states $b$ by $V^\pi(b) = E_b[\sum_{t=0}^\infty \gamma^t r_t]$, where $0<\gamma<1$ is the discount factor and $r_t$ is the reward received at the $t$th step in the future. Intuitively, it is the expected discounted reward the agent can expect to receive starting from belief state $b$ if it behaves according to policy $\pi$. An policy $\pi^*$ is *optimal* if $V^{\pi^*}(b) \geq V^\pi(b)$ for all $b$ and all other policies $\pi$. The value function of an optimal policy is called the *optimal value function* and is usually denoted by $V^*$.

Policies for POMDPs can be found through *value iteration* (Bellman 1957). Value iteration begins with an arbitrary initial function $V_0^*(b)$ and improves it by using the following equation

$$V_t^*(b) = max_a[r(b, a) + \gamma \sum_{o_+} P(o_+|b, a)V_{t-1}^*(b_+)], \quad (3)$$

where $P(o_+|b, a) = \sum_{s,s_+} P(s_+, o_+|s, a)b(s)$, and $b_+$ is a shorthand for $b_+(.|b, a, o_+)$. If $V_0^*=0$, $V_t^*$ is called the *t-step optimal value function*.

It is well known that when the *Bellman residual* $max_{b\in\mathcal{B}}|V_t^*(b) - V_{t-1}^*(b)|$ becomes small, $V_t^*$ is close to $V^*$ and the *greedy policy based on* $V_t^*$

$$\pi(b) = arg\ max_a[r(b, a) + \gamma \sum_{o_+} P(o_+|b, a)V_t^*(b_+)] \quad (4)$$

is a good approximation of the optimal policy (e.g. Puterman 1990).

Since there are uncountably infinite many belief states, value iteration cannot to carried out explicitly. Fortunately, it can be carried out implicitly due to the piecewise linearity of the $t$-step optimal value function (Sondik 1971). More specifically, there exists a list $\mathcal{V}_t$ of function of $s$, usually referred to simply as vectors, such that for any belief state

$$V_t^*(b) = max_{V\in\mathcal{V}_t} \sum_s V(s)b(s). \quad (5)$$

Exact methods for solving POMDPs (Monahan 1992, Eagle 1984, and Larke 1991 (see White 1991), Sondik 1971, Cheng 1988, Cassandra et al 1994) attempt to find a minimum list of vectors that satisfies the above equation. Unfortunately, even the most efficient algorithm can only solve POMDPs with no more than twenty states and fifteen observations exactly (Littman et al 1995, Cassandra et al 1997). Approximation is a must for real-world problems.

Most previous approximate methods (e.g. Cheng 1988, Lovejoy 1991b, and Parr and Russell 1995) attempt to find a list of vectors that satisfies equation (5) approximately. This paper proposes to approximate POMDPs themselves by others that have more informative observations and hence are easier to solve.

## 4 PROBLEM CHARACTERISTICS AND APPROXIMATIONS

We make the following assumption about problem characteristics. Even though in a POMDP $\mathcal{M}$ the agent does not know the true state of the world, he often has a good idea about it. See the introduction for justifications of this assumption.

Consider another POMDP $\mathcal{M}'$ which is the same as $\mathcal{M}$ except that in addition to the observation made by itself, the agent also receives a report from an oracle who knows the true state of the world. The oracle does not report the true state itself. Rather he selects, from a list of candidate regions, a region that contains the true state and report that region.

More information is available to the agent in $\mathcal{M}'$ than in $\mathcal{M}$; extra information is provided by the oracle. When the agent already has a good idea about the true state of the world, the oracle does not provides much extra information even when the candidate regions are small. In such a case, $\mathcal{M}'$ is a good approximation of $\mathcal{M}$.

In $\mathcal{M}'$, the agent knows for sure that the true state of the world is in the region reported by the oracle. For this reason, we say that it is *region observable*. The region observable POMDP $\mathcal{M}'$ can be much easier to solve than $\mathcal{M}$ when the candidate regions are small. For example, if the oracle is allowed to report only singleton regions, then he actually reports the true state of the world and hence $\mathcal{M}'$ is an MDP. MDPs are much easier the solve than POMDPs.

We now set out to make the idea more concrete. Let us begin with the concept of region systems.

### 4.1 Region Systems

A *region* is simply a subset of states of the world. A *region system* is a collection of regions such that no region is a subset of other regions in the collection and the union of all regions equals the set of all possible states of the world. We shall use $R$ to denote a region and $\mathcal{R}$ to denote a region system. Region systems are to be used to restrict the regions that the oracle can choose to report.

There are many possible ways to construct a region system. A natural way is to create a region for each state by including its "nearby" states. Let us make this more precise. Each action has an intended effect. The intended effect of move-forward, for instance, is to move one step forward. We say a state $s$ is *ideally reachable in one step* from another state $s'$ if there is an action whose intended effect is, when the world is currently in state $s'$, to take the world into state $s$. A state $s_k$ is *ideally reachable in $k$ steps* from another state $s_0$ if there are state $s_1, \ldots, s_{k-1}$ such that $s_{i+1}$ is ideally reachable from $s_i$ in one step for all $0 \leq i \leq k-1$. Any state is ideally reachable from itself in 0 step.

For any non-negative integer $k$, the *radius-$k$ region centered at a state $s$* consists of states that are ideally reachable from $s$ in $k$ or less steps. A *radius-$k$ region system* is the one obtained by creating a radius-$k$ region for each state and then removing, one after another, regions that are subsets of others.

When $k$ is 0, the radius-$k$ region system consists of singleton regions. On the other hand, if there is a $k$ such that any state is ideally reachable from any other state in $k$ or less steps, then there is only one region in the radius-$k$ region system, which is the set of all possible states.

### 4.2 Region Observable POMDPs

To complete the definition of the region observable POMDP $\mathcal{M}'$, assume a region system has been given and the oracle is allowed to choose region only from the system. This subsection discusses how the oracle should choose regions from the system. The main issue is to minimizes the amount of extra information.

To provide as little extra information as possible, the oracle should consider what the agent already knows. However, he cannot take the entire history of past actions and observations into account because if he did, $\mathcal{M}'$ would not be a POMDP. We suggest the following rule.

For any non-negative function $f(s)$ of $s$ and any region $R$, we call the quantity $supp(f, R) = \sum_{s \in R} f(s) / \sum_{s \in S} f(s)$ the *degree of support* of $f$ by $R$. If $R$ supports $f$ to degree 1, we say that $R$ *fully supports* $f$.

Let $s_-$ be the previous true state of the world, $a_-$ be the previous action, and $o$ be the current observation. The oracle should choose, among all the regions in $\mathcal{R}$ that contain the true state of the world, one that supports the function $P(s, o|s_-, a_-)$ of $s$ to the maximum degree. Where there is more than one such regions, choose the one that comes first in a predetermined ordering among the regions.

Here are the intuitions. If the previous world state $s_-$ were known to the agent, then his current belief state $b(s)$ would be proportional to $P(s, o|s_-, a_-)$. In this case, the rule minimizes extra information in the sense that it supports the current belief state to the maximum degree. Also if the current observation is informative enough, being a landmark for instance, to ensure that the world state is in a certain region, then region chosen using the rule fully supports the current belief state. In such a case, no extra information is provided.

We do not claim that the rule described above is optimal. Finding a rule that minimize extra information is still an open problem.

The probability $P(R|s, o, s_-, a_-)$ of a region $R$ being chosen under the above scheme is given by

$$P(R|s, o, s_-, a_-) = \begin{cases} 1 & \text{if } R \text{ is the first region s.t. } s \in R \\ & \text{and for any other region } R' \\ & \sum_{s' \in R} P(s', o|s_-, a_-) \geq \\ & \sum_{s' \in R'} P(s', o|s_-, a_-) \\ 0 & \text{otherwise.} \end{cases}$$

The region observable POMDP $\mathcal{M}'$ differs from the original POMDP $\mathcal{M}$ only in terms of observation; in addition to the observation $o$ made by himself, the agent also receives a report $R$ from the oracle. We shall denote an observation in $\mathcal{M}'$ by $z$ and write $z = (o, R)$. Observation model of $\mathcal{M}'$ is given by

$$P(z|s, a_-, s_-) = P(o, R|s, a_-, s_-) = P(o|s, a_-) P(R|s, o, s_-, a_-).$$

### 4.3 Solving Region Observable POMDPs

For any region $R$, let $\mathcal{B}_R$ be the set of belief states that are fully supported by $R$. For any region system $\mathcal{R}$, let $\mathcal{B}_\mathcal{R} = \cup_{R \in \mathcal{R}} \mathcal{B}_R$.

Let $\mathcal{R}$ be the region system underlying the region observable POMDP $\mathcal{M}'$. It is easy to see that no matter what the current belief state $b$ is, the next belief state $b_+$ must be in $\mathcal{B}_\mathcal{R}$. We assume that in $\mathcal{M}'$ the initial belief state is in $\mathcal{B}_\mathcal{R}$. Then all possible belief states the agent might have are in $\mathcal{B}_\mathcal{R}$. This implies that policies for $\mathcal{M}'$ need only be defined over $\mathcal{B}_\mathcal{R}$ and value iteration for $\mathcal{M}'$ can restricted to the subset $\mathcal{B}_\mathcal{R}$ of $\mathcal{B}$.

Restricting value iteration for $\mathcal{M}'$ to $\mathcal{B}_\mathcal{R}$ implies that the $t$-step optimal value function $U_t^*$ of $\mathcal{M}'$ is defined only over $\mathcal{B}_\mathcal{R}$ and the Bellman residual is now $max_{b \in \mathcal{B}_\mathcal{R}} |U_t^*(b) - U_{t-1}^*(b)|$.

Like value iteration, restricted value iteration can be carried out implicitly. Due to region observability, restricted implicit value iteration in $\mathcal{M}'$ can be done more efficiently than implicit value iteration in $\mathcal{M}$. See Zhang and Liu (1996) for details.

Implicit restrict value iteration gives us a vectors, which will be henceforth denoted by $\mathcal{U}_t$. It represents the $t$-step optimal value function $U_t^*(b)$ of $\mathcal{M}'$ in the sense that $U_t^*(b) = max_{V \in \mathcal{U}_t} \sum_s b(s) V(s)$ for any $b \in \mathcal{B}_\mathcal{R}$. The greedy policy for $\mathcal{M}'$ based on $U_t^*$ is as follows: for any $b \in \mathcal{B}_\mathcal{R}$

$$\pi'(b) = arg\ max_a [r(b,a) + \gamma \sum_{z_+} P(z_+|b,a) U_t^*(b_+)], \quad (6)$$

where $z_+$ stands for observation of the next time point and $b_+$ is a shorthand for the next belief state $b_+(.|b, a, z_+)$.

## 5   POLICY FOR THE ORIGINAL POMDP

Suppose we have solved the region observable POMDP $\mathcal{M}'$. The next step is to construct a policy $\pi$ for the original POMDP $\mathcal{M}$ based on the solution for $\mathcal{M}'$.

Even though it is our assumption that in the original POMDP $\mathcal{M}$ the agent has a good idea about the state of the world at all time, there is no guarantee that its belief state will always be in $\mathcal{B}_\mathcal{R}$. There is no oracle in $\mathcal{M}$. A policy should prescribes actions for belief states in $\mathcal{B}_\mathcal{R}$ as well as for belief states outside $\mathcal{B}_\mathcal{R}$. An issue here is that the policy $\pi'$ for $\mathcal{M}'$ is defined only for belief states in $\mathcal{B}_\mathcal{R}$. Fortunately, $\pi'$ can be naturally extended to the entire belief space by ignoring the constraint $b \in \mathcal{B}_\mathcal{R}$ in equation (6). We hence define an policy $\pi$ for $\mathcal{M}$ as follows: for any $b \in \mathcal{B}$,

$$\pi(b) = arg\ max_a [r(b,a) + \gamma \sum_{z_+} P(z_+|b,a) U_t^*(b_+)]. \quad (7)$$

Let $k$ be the radius of the region system underlying $\mathcal{M}'$. The policy $\pi$ for $\mathcal{M}$ given above will be referred to as the *radius-k approximate policy* for $\mathcal{M}$. The entire process of obtaining the policy, including the construction and solving of the region observable POMDP $\mathcal{M}'$, will be referred to as *region-based approximation*.

It is worthwhile to compare this equation with equation (4). In equation (4), there are two terms on the right hand side. The first term is the immediate reward for taking action $a$ and the second term is the discounted future reward the agent can expect to receive if it behaves optimally. Their sum is the total expected reward for taking action $a$. The action with the highest total reward is chosen.

The second term is difficult to obtain. In essence, equation (7) approximates the second term using the optimal expected future reward the agent can receive with the help of the oracle, which is easier to compute.

It should be emphasized that the presence of the oracle is assumed only in the process of computing the radius-k approximate policy. The oracle is not present when executing the policy.

## 6   QUALITY OF APPROXIMATION AND SIMULATION

In general, the quality of an approximate policy $\pi$ is measured by the distance between the optimal value function $V^*(b)$ and the value function $V^\pi(b)$ of $\pi$. This measurement does not consider what the agent might know about the initial state of the world. As such, it is not appropriate for a policy obtained through region-based approximation. One cannot expect such a policy be of good quality if the agent is very uncertain about the initial state of the world because it is obtained under the assumption that the agent has a good idea about the state of the world at all time.

This section describes a scheme for determining the quality of an approximate policy in cases where the agent knows the initial state of the world with certainty. The scheme can be generalized to cases where there is a small amount of uncertainty about the initial state; for example, cases where the initial state is known to be in some small region.

The agent might need to reach the goal from different initial states at different times. Let $P(s)$ be the frequency it will start from state $s$[1]. The quality of an approximate policy $\pi$ can be measured by $\sum_s |V^*(s) - V^\pi(s)| P(s)$, where $V^*(s)$ and $V^\pi$ denote the rewards the agent can expect to receive starting from state $s$ if it behaves optimally or according to $\pi$ respectively.

By definition $V^*(s) \geq V^\pi(s)$ for all $s$. Let $U^*$ be the optimal value function of the region observable POMDP $\mathcal{M}'$. Since more information is available to the agent in $\mathcal{M}'$, $U^*(s) \geq V^*(s)$ for all $s$. Therefore, $\sum_s [U^*(s) - V^\pi(s)] P(s)$ is an upper bound on $\sum_s [V^*(s) - V^\pi(s)] P(s)$.

Let $\pi'$ be the policy for $\mathcal{M}'$ given by (6). When the Bellman residual is small, $\pi'$ is close to optimal for $\mathcal{M}'$ and the value function $V^{\pi'}$ of $\pi'$ is close to $U^*$. Consequently, $\sum_s [V^{\pi'}(s) - V^\pi(s)] P(s)$ is an upper bound on $\sum_s [V^*(s) - V^\pi(s)] P(s)$ when the Bellman residual is small enough.

---

[1] This is not to be confused with the initial belief state $P_0$.

One way to estimate the quantity $\sum_s [V^{\pi'}(s) - V^\pi(s)] P(s)$ is to conduct a large number of simulation trials. In each trial, an initial state is randomly generated according to $P(s)$. The agent is informed of the initial state. Simulation takes place in both $\mathcal{M}$ and $\mathcal{M}'$. In $\mathcal{M}$, the agent chooses, at each step, an action using $\pi$ based on the its current belief state. The action is passed to a simulator which randomly generates the next state of the world and the next observation according to the transition and observation probabilities. The observation (but not the state) is passed to the agent, who updates its belief state and chooses the next action. And so on and so forth. The trial terminates when the agent chooses the action declare-goal or a maximum number of steps is reached. Simulation in $\mathcal{M}'$ takes place in a similar manner except that the observations and the observation probabilities are different and actions are chosen using $\pi'$.

If the goal is correctly declared at the end of a trial, the agent receives a reward of the amount $\gamma^n$, where $n$ is the number of steps. Otherwise, the agent receive no reward. The quantity $\sum_s [V^{\pi'}(s) - V^\pi(s)] P(s)$ can be estimated using the difference between the average reward received in the trials for $\mathcal{M}'$ and the average reward received in the trials for $\mathcal{M}$.

## 7 TRADEOFF BETWEEN QUALITY OF APPROXIMATION AND COMPLEXITY

Intuitively, the larger the radius of the region system, the less the amount of extra information the oracle provides. Hence the closer $\mathcal{M}'$ is to $\mathcal{M}$ and the narrower the gap between $\sum_s V^{\pi'}(s) P(s)$ and $\sum_s V^\pi(s) P(s)$. Although we have not theoretically proved this, empirical results (see the next section) do suggest that $\sum_s V^\pi(s) P(s)$ increases with the radius of the region system while $\sum_s V^{\pi'}(s) P(s)$ decreases with it. At the extreme case when there is one region in the region system that contains all the possible states of the world, $\mathcal{M}$ and $\mathcal{M}'$ are identical and hence so are $\sum_s V^{\pi'}(s) P(s)$ and $\sum_s V^\pi(s) P(s)$.

Those discussions lead to the following scheme for making the tradeoff between complexity and quality. Start with the radius-0 region system and increases the radius gradually until the quantity $\sum_s [V^{\pi'}(s) - V_\pi(s)] P(s)$ becomes sufficiently small or the region observable POMDP $\mathcal{M}'$ becomes untractable.

## 8 SIMULATION EXPERIMENTS

Simulation experiments have been carried out to show that (1) quality of approximation increased with radius of region system and (2) where there is not much uncertainty, a POMDP can be accurately approximated by a region-observable POMDP that can be solved exactly. This section reports on the experiments.

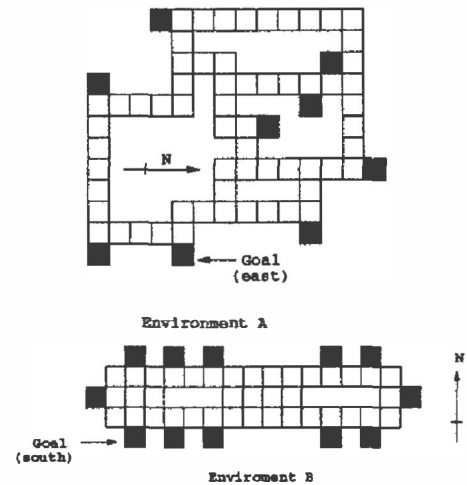

Figure 1: Synthetic Office Environments.

### 8.1 Synthetic Office Environments

Our experiments were carried using two synthetic office environments borrowed from Cassandra *et al* (1996) with some minor modifications. Layouts of the environments are shown in Figure 1, where squares represent locations. Each location is represented as four states in the POMDP model, one for each orientation. The dark locations are rooms connected to corridors by doorways.

In each environment, a robot needs to reach the goal location with the correct orientation. At each step, the robot can execute one of the following actions: move-forward, turn-left, turn-right, and declare-goal. The two sets of action models given in the following table were used.

| Action | Standard outcomes | Noisy outcomes |
|---|---|---|
| move-forward | N(0.11), F(0.88), F-F(0.01) | N(0.2), F(0.7), F-F(0.1) |
| turn-left | N(0.05), L(0.9), L-L(0.05) | N(0.15), L(0.7), L-L(0.15) |
| turn-right | N(0.05), R(0.9), R-R(0.05) | N(0.15), R(0.7), R-R(0.15) |
| declare-goal | N(1.0) | N(1.0) |

For the action move-forward, the term F-F (0.01) means that with probability 0.01 the robot actually moves two steps forward. The other terms are to be interpreted similarly. If an outcome cannot occur in a certain state of the world, then the robot is left in the last state before the impossible outcome.

In each state, the robot is able to perceive in each of three nominal directions (front, left, and right) whether there is a doorway, wall, open, or it is undetermined. The following two sets of observation models were used:

| Actual case | Standard observations | Noisy observations |
|---|---|---|
| wall | wall (0.90), open (0.04), doorway (0.04), undetermined (0.02) | wall (0.70), open (0.19), doorway (0.09), undetermined (0.02) |
| open | wall (0.02), open (0.90), doorway (0.06), undetermined (0.02) | wall (0.19), open (0.70), doorway (0.09), undetermined (0.02) |
| doorway | wall (0.15), open (0.15), doorway (0.69), undetermined (0.01) | wall (0.15), open (0.15), doorway (0.69), undetermined (0.01) |

### 8.2 Complexity of Solving the POMDPs

One of the POMDPs have 280 possible states while the other has 200. They both have 64 possible observations and 4 possible actions. Since the largest POMDPs that researchers have been able to solve exactly so far have less than 20 states and 15 observations, it is safe to say no existing exact algorithms can solve those two POMDPs.

We were be able to solve the radius-0 and radius-1 approximations (region observable POMDPs) of the two POMDPs on a SUN SPARC20 computer. The threshold for the Bellman residual was set at 0.001 and the discount factor at 0.99. The amounts of time it took in CPU seconds are collected in the following table.

| Environ -ment | Standard models | | Noisy models | |
|---|---|---|---|---|
| | Radius-0 | Radius-1 | Radius-0 | Radius-1 |
| A | 1.26 | 4135 | 1.35 | 6571 |
| B | 0.61 | 3083 | 0.72 | 5426 |

We see that the radius-1 approximations took much longer time to solve than the radius-0 approximations. Also notice that the region observable POMDPs with noisy action and observation models took more time to solve that those with the standard models.

We were unable to solve the radius-2 approximations. Other approximation techniques need to be incorporated in order to solve the approximations based on region systems with radius larger than or equal to 2.

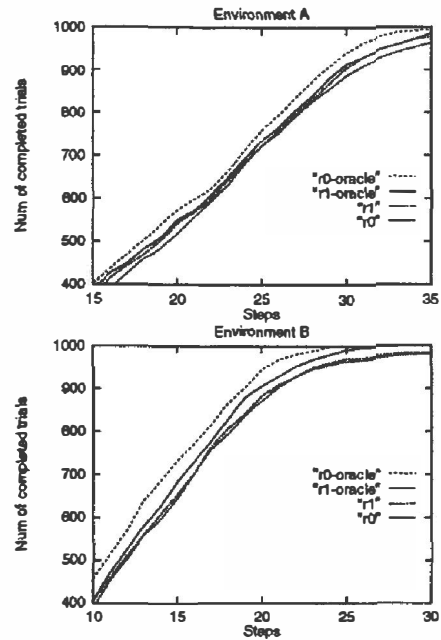

Figure 2: Experiments with standard action and noisy models. The POMDPs are accurately approximated by region observable POMDPs with radius zero or one.

### 8.3 Quality of Approximation for Standard Models

To determine the quality of the radius-0 and radius-1 approximate policies for the POMDPs with standard action and observation models, 1000 simulation trials were conducted using the scheme described in Section 6. It was assumed that the agent is equally likely to start from any state. Instead of the average reward over the trials, the performance of the agent is summarized by the distribution of the numbers of steps it took to successfully complete the trials, i.e. by a function $g(n)$ of steps $n$, where for each $n$, $g(n)$ is the number of trials where the goal was reached and declared in $n$ or less steps. The average reward over the trials can be computed by $\sum_{n=0}^{100} \gamma^n (g(n)-g(n-1))/1000$. We choose the function $g(n)$ instead of the average reward because it is more informative than the latter.

Simulation results are shown in Figure 2. The curves r0-oracle, for instance, represent the g-functions for simulations in the radius-0 region observable POMDPs (i.e. with the help of the oracle) using their optimal policies. In contrast, the curves r0 represent the g-functions for simulations in the original POMDPs (without the help of the oracle) using radius-0 approximate policies. For readability, only top portions of the g-functions are shown.

We see that the gap between r0-oracle and r0 is quite small in both cases. This indicates that the radius-0 region observable POMDPs (MDPs) are quite accurate approximations of the original POMDPs. The radius-0 approximate policies are close to optimal for the original POMDPs.

The gaps between the curves r1-oracle and r1 are even narrower. For environment A, there is essentially no gap. Also notice that the curves r1 lie above r0 and the curves r1-oracle lie below r0-oracle. Those support our claim that quality of approximation increases with radius of region system.

There is a couple other facts worth mentioning. The gaps are larger in environment B than in environment A. This is because environment B is more symmetric and consequently observations are less effective in disambiguating uncertainty in the agent's belief about the state of the world.

There were a few failures in environment A even with the presence of the oracle (curve r1-oracle). The failures occurred due to uncertainties in the actions models: The agent was one step away from the goal and had an very good idea about the state of the world. An action towards the goal was taken and afterwards the agent believed strongly that the world is in the goal state. However, the action failed to effect any movement and the orcale's report did point this out[2]. So a failure.

### 8.4 Quality of Approximation for Noisy Models

One thousand trials were also conducted for the POMDPs with noisy action and observation models. Results are shown in Figure 3.

We see that the gaps between r1-oracle and r1 is significantly narrower than the gaps between r0-oracle and r0, especially for environment A. The curves r1 lie above the curves r0 and the curves r1-oracle lie below r0-oracle. Again, those support our claim that quality of approximation increases with radius of region system.

As far as absolute quality of approximation is concerned, the radius-0 POMDPs are obviously very poor approximations of the original POMDPs since the gaps gaps between the curves r0-oracle and r0 are very wide. For Environment A, the radius-1 approximation is fairly accurate. However, the radius-1 approximation remains poor for environment B. The radius of region system needs to be increased. Unfortunately, increasing the radius beyond 1 renders it computationally impossible to solve the region observable

---
[2]The oracle reported a region that contains both the goal and the actual state.

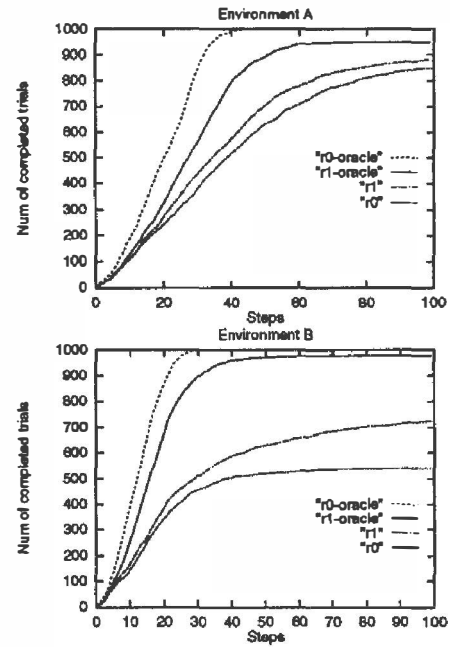

Figure 3: Experiments with noisy action and noisy models. The POMDPs are not accurately approximated by region observable POMDPs with radius zero or one.

POMDPs exactly.

Tracing through the trials, we learned some interesting facts. In environment B, the agent, under the guidance of the radius-1 approximate policy, was able to quickly get to the neighborhood of the goal even when starting from far way. The fact that the environment around the goal is highly symmetric was the cause of the poor performance. Often the agent was not able to determine whether it was at the goal location (room), or in the opposite room, or in the left most room, or in the room to the right of the goal location. The performance would be close to optimal if the goal location had some distinct features.

In environment A, the agent, again under the guidance of the radius-1 approximate policy, was able to reach and declare the goal successfully once it got to the neighborhood. However, it often took many unnecessarily steps before reaching the neighborhood due to the undesirable effects of the turning actions. Take the lower left corner as an example. When the agent reached the corner from above, it was facing downward. The agent executed the action turn_left. Fifteen percent of the time, it ended up facing upward instead of to the right — the desired direction. The agent then decided to move-forward, thinking that it was approaching the goal. But it was actually moving upward and did not realize this until a few steps later.

The agent would perform much better there were informative landmarks around the corners.

## 9 CONCLUSIONS

We propose to approximate a POMDP by using a region observable POMDP. The region observable POMDP has more informative observations and hence is easier to solve. A method for determining the quality of approximation is also described, which allows one to make the tradeoff between quality of approximation and computational complexity by starting with a coarse approximation and refining it gradually. Simulation experiments have shown that when there is not much uncertainty in the effects of actions and observations are informative, a POMDP can be accurately to approximated by a region observable POMDP that can be solved exactly. However, this becomes infeasible as the degree of uncertainty increases. Other approximate methods need to be incorporated in order to solve region observable POMDPs whose radiuses are not small.

### Acknowledgement

Research was supported by Hong Kong Research Council under grants HKUST 658/95E and Hong Kong University of Science and Technology under grant DAG96/97.EG01(RI).